\documentclass[twocolumn]{article}
\usepackage{graphicx} 
\usepackage{cite}
\usepackage{amsmath} 

\title{Optimizing Low-Speed Autonomous Driving: A Reinforcement Learning Approach to Route Stability and Maximum Speed}
\author{Benny Bao-Sheng Li, Elena Wu, Hins Shao-Xuan Yang, Nicky Yao-Jin Liang}
\date{October 2024}

\begin{document}

\maketitle

\section*{ABSTRACT}
\addcontentsline{toc}{section}{Abstract}
Autonomous driving has garnered significant attention in recent years, especially in optimizing vehicle performance under varying conditions. This paper addresses the challenge of maintaining maximum speed stability in low-speed autonomous driving while following a predefined route. Leveraging reinforcement learning (RL), we propose a novel approach to optimize driving policies that enable the vehicle to achieve near-maximum speed without compromising on safety or route accuracy, even in low-speed scenarios.

\paragraph{}
Our method uses RL to dynamically adjust the vehicle’s behavior based on real-time conditions, learning from the environment to make decisions that balance speed and stability. The proposed framework includes state representation, action selection, and reward function design that are specifically tailored for low-speed navigation. Extensive simulations demonstrate the model’s ability to achieve significant improvements in speed and route-following accuracy compared to traditional control methods.

\paragraph{}
These findings suggest that RL can be an effective solution for enhancing the efficiency of autonomous vehicles in low-speed conditions, paving the way for smoother and more reliable autonomous driving experiences.

\newpage
\section{INTRODUCTION}
Reinforcement Learning (RL) has become a powerful approach for addressing complex decision-making challenges in autonomous systems, particularly in low-speed scenarios. Unlike high-speed driving, low-speed environments demand high precision, safety, and stability \cite{schulman2015trust} due to dynamic obstacles and confined spaces. This paper explores several applications of RL in low-speed contexts, demonstrating its potential to enhance performance in various tasks.

\begin{enumerate}
  \item Autonomous Parking Systems: RL optimizes vehicle parking maneuvers in tight spaces, allowing vehicles to learn adjustments in steering and speed for efficient parking while avoiding obstacles.
  \item Low-Speed Car Following: In congested traffic, RL enables vehicles to dynamically adjust their speed based on the lead vehicle's behavior, improving safety and comfort.
  \item Urban Low-Speed Navigation: RL facilitates autonomous navigation in urban environments, allowing vehicles to adapt to changing conditions and outperform traditional control methods.
  \item Delivery and Shuttle Vehicles: RL enhances the navigation of autonomous delivery robots and shuttles in complex environments, optimizing efficiency and safety at low speeds.
  \item Obstacle Traversal in Off-Road Vehicles: RL improves control strategies for off-road vehicles traversing uneven terrain, maintaining stability while avoiding damage.
  \item Low-Speed Drone Navigation: For drones operating in confined spaces, RL enhances obstacle avoidance, ensuring safe navigation where GPS signals are weak.
  \item Service Robot Navigation: In indoor settings, RL optimizes the movement of service robots through dynamic environments, ensuring safety around humans.
\end{enumerate}

These applications highlight RL’s ability to meet the unique challenges of low-speed navigation. This paper focuses on using RL to maximize speed stability in low-speed autonomous driving while following predefined routes, presenting a novel approach to optimize both speed and safety in constrained environments.

\section{BACKGROUNDS}
One of the critical challenges in autonomous driving is ensuring that vehicles can follow predefined routes at maximum speed while maintaining stability, especially in low-speed conditions. This problem becomes more complex in real-world scenarios where safety, precision, and adaptability are paramount.

To test and validate our proposed RL-based approach, we conduct experiments using the AWS DeepRacer platform. AWS DeepRacer offers a simulated environment where autonomous models can be trained and tested in real-time, providing a practical and scalable solution for RL experimentation. Specifically, we leverage the AWS DeepRacer Student League\cite{awsdeepracer} to rigorously evaluate our algorithm's performance. Our focus is on optimizing for the shortest path, maximum speed, and stability, aiming to demonstrate that our RL approach can outperform traditional control methods in low-speed, complex environments.

Through these experiments, we aim to provide a robust validation of our approach, showcasing its potential application in real-world autonomous driving systems.

\section{METHODS}
\subsection{Speed Control at Maximum Limit}
The primary objective of our algorithm is to ensure that the vehicle consistently operates at its maximum allowable speed of 1 m/s, regardless of the driving conditions. We already know that the speed limit is constrained to 1 m/s, so the challenge is to develop a control mechanism that allows the vehicle to sustain this top speed in all scenarios, such as navigating curves, straight paths, and while avoiding obstacles. The algorithm dynamically adjusts the vehicle’s acceleration and deceleration to keep it at this speed limit, ensuring that any deviations due to environmental factors are promptly corrected. This speed control is critical for optimizing performance while adhering to safety requirements.

Quadratic and Exponential Rewards in Robotics\cite{ng1999reward}: Nonlinear reward functions such as quadratic or exponential penalties are commonly used in robotics for tasks like navigation and speed control. These functions help to penalize large deviations more severely, which can be crucial for maintaining stability at high or low speeds, nonlinear reward functions have been applied successfully in robotics.

Quadratic and exponential reward functions are commonly used in reinforcement learning (RL) to optimize speed control in robotic and autonomous systems. These nonlinear rewards help balance achieving maximum speed while ensuring safety and stability.

An exponential reward function penalizes deviations from the target speed even more aggressively, which can be useful in environments where maintaining speed is crucial for performance.
\begin{equation}
R(s,a) = e^{-\alpha |v_{\text{target}} - v_{\text{actual}}|}
\end{equation}

\textbf{Where:}
\begin{itemize}
    \item \(\alpha\) is a scaling factor controlling how steep the penalty is.
    \item \(v_{\text{target}}\) is the desired speed (e.g., 1 m/s).
    \item \(v_{\text{actual}}\) is the actual speed.
\end{itemize}

\begin{figure}[h]
    \centering
    \includegraphics[width=0.87\linewidth]{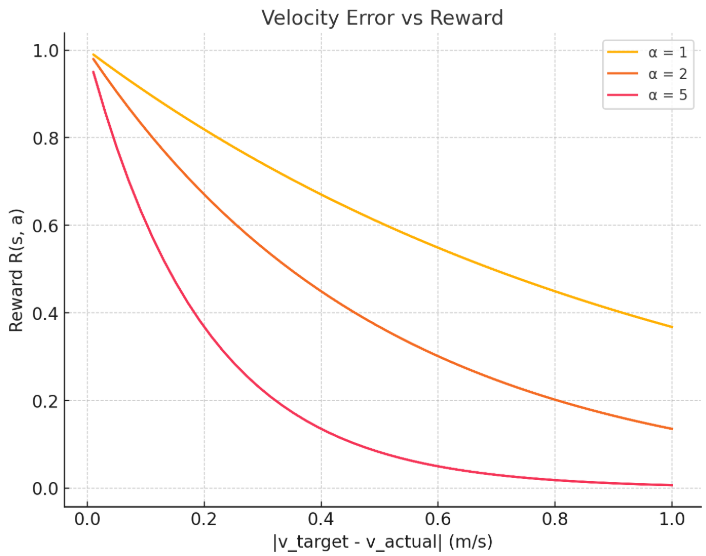}
    \caption{Speed}
    \label{fig:enter-label}
\end{figure}

\begin{itemize}
    \item The x-axis represents the velocity error \( \left| v_{\text{target}} - v_{\text{actual}} \right| \), which is the absolute difference between the target and actual speeds.
    \item The y-axis represents the reward \( R(s, a) = e^{-\alpha \left| v_{\text{target}} - v_{\text{actual}} \right|} \), which decreases as the error increases.
    \item Different values of \( \alpha \) control the rate at which the reward decays. A larger \( \alpha \) leads to a faster decay of the reward as the velocity error increases.
\end{itemize}

\begin{figure}[h]
    \centering
    \includegraphics[width=0.87\linewidth]{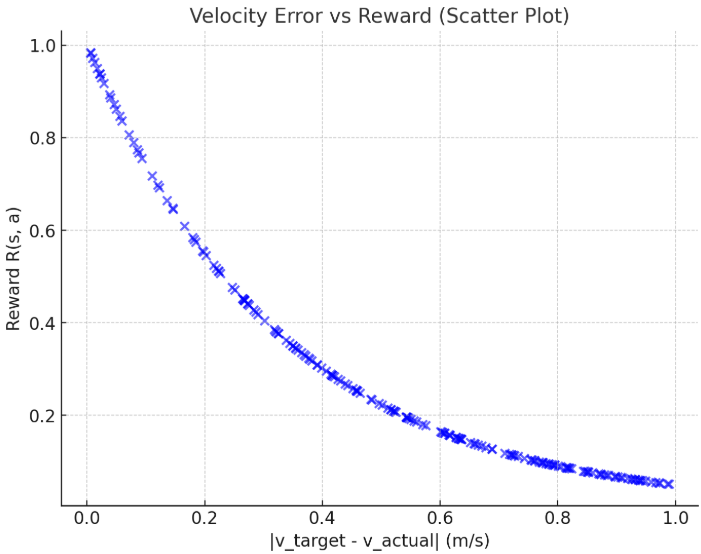}
    \caption{Speed}
    \label{fig:enter-label}
\end{figure}

This scatter plot shows the relationship between the velocity error \( \left| v_{\text{target}} - v_{\text{actual}} \right| \) and the corresponding reward \( R(s, a) \) for a fixed \( \alpha = 3 \):
\begin{itemize}
    \item Each point represents an individual sample where the actual velocity is generated randomly.
    \item As the velocity error increases, the reward decreases exponentially.
    \item The scatter plot effectively visualizes the distribution and spread of the rewards based on the velocity error.
\end{itemize}

Figure~X presents the relationship between the velocity error \( \left| v_{\text{target}} - v_{\text{actual}} \right| \) and the corresponding reward \( R(s, a) \) under varying parameter settings. Specifically, we trained the model using different values of \( \alpha \), where \( \alpha \) controls the rate of reward decay as the velocity error increases. The scatter plot illustrates the distribution of rewards for randomly sampled actual velocities (\( v_{\text{actual}} \)) within the range [0, 1] m/s.

From the results, it is evident that as the velocity error increases, the reward decays exponentially, adhering to the formula:
\[
R(s, a) = e^{-\alpha \left| v_{\text{target}} - v_{\text{actual}} \right|}.
\]
Larger values of \( \alpha \) lead to a sharper decline in the reward, highlighting a stricter penalty for deviations from the target velocity. This observation demonstrates the sensitivity of the reward function to hyperparameter tuning and its influence on the model's learning behavior.

\bigskip
Exponential rewards are more sensitive to deviations, making them ideal for strict control tasks where even small speed variations can cause significant performance problems.

\subsection{Progress-Based Rewards in Autonomous Driving \cite{shalev2016deep}}
In autonomous driving, progress-based rewards are often used to incentivize the vehicle to follow a specific route or track as efficiently as possible. The idea is that the agent (vehicle) receives a reward based on how much progress it makes toward completing a lap, reaching a goal, or following a path. Incorporating a ratio of progress made to distance traveled is a way to ensure that the agent does not take unnecessarily long or inefficient routes.

Progress-Based Reward Using Incremental Distance  \cite{lavalle2006planning} :
\begin{equation}
R_t = \frac{\Delta \text{Progress}}{\Delta L}
\end{equation}

\textbf{Where:}
\begin{itemize}
    \item \( R_t \) is the reward at time step \( t \).
    \item \( \Delta \text{Progress} = \text{Progress}_t - \text{Progress}_{t-1} \) is the change in progress along the route between consecutive time steps.
    \item \( \Delta L = L_t - L_{t-1} \) is the incremental distance traveled between the two time steps.
\end{itemize}

This design of the reward function \cite{sutton2018reinforcement} encourages the agent to maximize progress while minimizing unnecessary distance traveled, promoting more efficient movement.

The formula \( R_t = \frac{\Delta \text{Progress}}{\Delta L} \) provides a straightforward and interpretable metric for evaluating traversal efficiency in autonomous systems. However, numerical stability issues may arise under specific conditions, which can affect the reliability and robustness of the metric.

\paragraph{Key Challenges in Numerical Stability}
\begin{enumerate}
    \item Small-Denominator Problem: \begin{itemize}
            \item When the path increment \( \Delta L \) approaches zero, the value of \( R_t \) can become exceedingly large or undefined. This situation commonly occurs in scenarios involving sharp turns, low-speed maneuvers, or near-stationary conditions.
            \item Such instability not only distorts the evaluation metric but may also propagate errors into the learning process in reinforcement learning frameworks.
        \end{itemize}
    \item Sensitivity to Noise:
        \begin{itemize}
            \item Measurement errors or noise in position data can significantly affect \( \Delta \text{Progress} \) and \( \Delta L \) , particularly when these increments are small. Even minor inaccuracies can lead to disproportionate changes in the metric.
        \end{itemize}
    \item Discontinuous Behavior:
     \begin{itemize}
            \item In complex trajectories with frequent stops and starts, the metric can exhibit abrupt changes, leading to challenges in reward function continuity during training.
        \end{itemize}
\end{enumerate}

To address the instability caused by small path incremental (\(\Delta L \to 0\)), a regularization term \(\epsilon > 0\) is introduced to stabilize the metric. The modified formula is expressed as:

\[
R_t = \frac{\Delta \text{Progress}}{\Delta L + \epsilon},
\]

where \(\epsilon\) is a small positive constant. This adjustment ensures that the denominator never approaches zero, preventing large or undefined values for \(R_t\). Regularization improves the numerical stability of the metric while maintaining its interpretability.

The Parameter \(\epsilon\) is critical for achieving a balance between stability and sensitivity, which can be fixing value or the cure:
\begin{enumerate}
     \item Small \(\epsilon\):
        \begin{itemize}
            \item Retains high sensitivity to variations in \( \Delta L \).
            \item Risk of instability when \( \Delta L \)  is extremely small.
        \end{itemize}
     \item Large \(\epsilon\):
        \begin{itemize}
            \item Improves stability but reduces the metric's responsiveness to changes in \( \Delta L \).
        \end{itemize}
\end{enumerate}

An empirical or adaptive approach can be used to tune \(\epsilon\) depending on the application scenario:

\[
\epsilon = \alpha \cdot \text{mean}(\Delta L)
\]

where \(\alpha\) is a scaling factor that controls the sensitivity of the regularization term. By dynamically adjusting \(\epsilon\), the metric can maintain stability while remaining responsive to variations in path increments.

For the racing training it can be implemented dynamically based on trajectory characteristics:

\begin{enumerate}
     \item Context-Aware Regularization:
        \begin{itemize}
            \item Adjust \(\epsilon\) based on environmental factors, such as the path curvature \cite{frazzoli2002real} or speed profile.
        \end{itemize}
     \item Time-Decaying Regularization:
        \begin{itemize}
            \item \(\epsilon\) can be designed to decay over time. This allows the model to focus on finer details as training progresses. The time-decaying regularization is expressed as:
                \[
                \epsilon_t = \epsilon_0 \cdot e^{-\beta t},
                \]
                where \(\epsilon_0\) is the initial regularization value, \(\beta > 0\) is the decay rate, and \(t\) represents the training step. By reducing \(\epsilon\) over time, the metric transitions from emphasizing stability in the early stages to precision in the later stages of training.
        \end{itemize}
\end{enumerate}

For this improvement it can benefit the training result :
\begin{enumerate}
     \item Prevention of Instability:
        \begin{itemize}
            \item By ensuring that the denominator never approaches zero, the metric avoids producing exceedingly large or undefined values.
            \item This is particularly useful in scenarios involving near-zero path increments, such as during sharp turns or low-speed maneuvers.
        \end{itemize}
     \item Enhanced Robustness:
        \begin{itemize}
            \item The addition of \(\epsilon\) makes the metric less sensitive to small variations in \( \Delta L \), improving reliability under noisy conditions.
        \end{itemize}
    \item Continuity in Training:
        \begin{itemize}
            \item Regularization reduces abrupt changes in the reward function, facilitating smoother gradient updates during the learning process.
    \end{itemize}
\end{enumerate}

Final Experimental Validation to validate the effectiveness of smoothing via regularization, a comparative study can be conducted using:

\begin{itemize}
    \item \textbf{Unregularized Formula:}
    \[
    R_t = \frac{\Delta \text{Progress}}{\Delta L}.
    \]

    \item \textbf{Regularized Formula:}
    \[
    R_t = \frac{\Delta \text{Progress}}{\Delta L + \epsilon}.
    \]
\end{itemize}

Metrics such as training stability, reward consistency, and trajectory optimization efficiency can be evaluated to demonstrate the benefits of regularization.
\begin{figure}[h]
    \centering
    \includegraphics[width=1.0\linewidth]{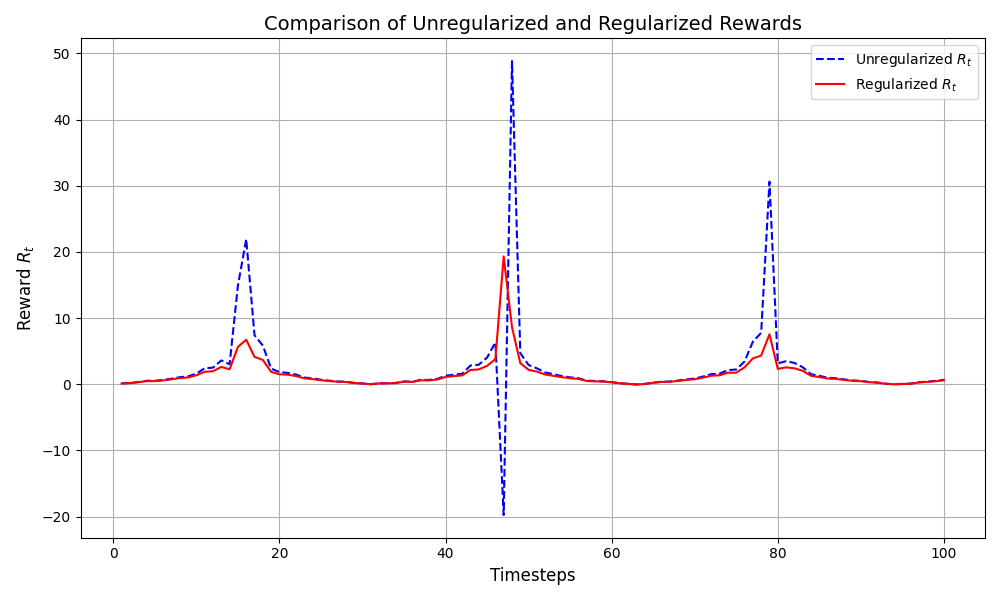}
    \caption{Progress-Based}
    \label{fig:enter-label}
\end{figure}

It demonstrates the differences between the unregularized and regularized reward formulas. The unregularized formula \( R_t = \frac{\Delta \text{Progress}}{\Delta L} \) is highly sensitive to small path increments (\(\Delta L\)), leading to large fluctuations and potential instability. 

In contrast, the regularized formula \( R_t = \frac{\Delta \text{Progress}}{\Delta L + \epsilon} \) smooths the reward function by introducing a regularization term \(\epsilon > 0\). This adjustment improves numerical stability and ensures that the reward values remain bounded, even in scenarios involving sharp turns, low-speed maneuvers, or near-stationary conditions. 

These results highlight the importance of incorporating regularization into the reward function to enhance robustness and reliability during training and evaluation.

\subsection{Smooth Steering Control in Autonomous Driving\cite{hanna2017minimum}}
A common approach to penalize excessive steering changes is by incorporating a term in the reward function that penalizes large deltas (changes) in the steering angle between two consecutive time steps. The penalty can either be proportional to the magnitude of the change or quadratic to heavily penalize larger changes. Here’s an example of a steering delta penalty.

\begin{equation}
R_{\text{steer}} = -k \cdot \left| \Delta \theta_{\text{steer}} \right|
\end{equation}

\textbf{Where:}
\begin{itemize}
    \item \( R_{\text{steer}} \) is the steering penalty applied to the overall reward.
    \item \( \Delta \theta_{\text{steer}} = \theta_{\text{steer},t} - \theta_{\text{steer},{t-1}} \) is the difference in steering angle between two consecutive time steps \( t \) and \( t-1 \).
    \item \( k \) \cite{sutton2018reinforcement} is a scaling factor that determines how much to penalize larger steering changes.
\end{itemize}

This reward component is designed to penalize excessive steering angle changes, encouraging smoother and more stable driving behavior.

\paragraph{Key Characteristics of the Formula:}
\begin{enumerate}
    \item Linear Penalization:
        \begin{itemize}
            \item The penalty is directly proportional to the absolute value of the steering change. Larger steering adjustments result in higher penalties.
        \end{itemize}
    \item Encouragement of Smooth Steering:
        \begin{itemize}
            \item By penalizing frequent or abrupt steering changes, the formula incentivizes the agent to follow trajectories with minimal oscillations, promoting smoother control dynamics.
    \end{itemize}
     \item Parameter Sensitivity  \(k\) :
        \begin{itemize}
            \item A higher value of \(k\) imposes a stricter penalty for steering changes, discouraging large steering adjustments. However, this can lead to overly cautious behavior, where the agent avoids necessary steering actions, particularly in scenarios requiring sharp turns.
            \item A lower value of \(k\) reduces the impact of the penalty, allowing the agent more flexibility in steering adjustments. While this promotes responsiveness, it can potentially result in unnecessary oscillations or instability in the trajectory.
    \end{itemize}
\end{enumerate}

The advantages of this formula in the racing training:
\begin{enumerate}
    \item Improved Vehicle Stability:
        \begin{itemize}
            \item Penalizing sharp steering changes helps to maintain vehicle stability, particularly at higher speeds where abrupt steering can lead to unsafe maneuvers.
        \end{itemize}
    \item Energy Efficiency:
        \begin{itemize}
            \item Smoother steering reduces unnecessary energy consumption, which is especially relevant in autonomous electric vehicles.
    \end{itemize}
     \item Trajectory Smoothness:
        \begin{itemize}
            \item The reward promotes smooth trajectories, improving both passenger comfort and adherence to the desired path.
    \end{itemize}
\end{enumerate}

Extend the formula and implement into the training and evaluate results:

In scenarios requiring sharp turns, the steering penalty may discourage necessary adjustments due to the increased cost associated with large steering angle changes. This can lead to suboptimal trajectory planning, particularly in environments with high curvature.

To mitigate this issue, the penalty can be weighted \cite{rajamani2011vehicle} by the curvature \cite{macadam2003understanding} of the path:
\[
R_\text{steer} = -k \cdot |\Delta \theta_\text{steer}| \cdot (1 - w_\text{curve}),
\]

where:
\begin{itemize}
    \item \( |\Delta \theta_\text{steer}| \) represents the absolute change in steering angle.
    \item \( k \) \cite{sutton2018reinforcement} is the proportionality constant controlling the penalty magnitude.
    \item \( w_\text{curve} \) is a weighting factor proportional to the curvature of the path. It increases in regions of high curvature, reducing the penalty to encourage necessary adjustments.
\end{itemize}

\( w_\text{curve} \) is important factor in the path, so it can use it as dynamic weight improve the formula:

\begin{enumerate}
    \item Dynamic Curvature Weighting:
        \begin{itemize}
            \item Implement a curvature-based weighting function \(w_\text{curve} = \min(\gamma \cdot \text{curvature}, 1)\), where \(\gamma\) is a scaling factor. This ensures the penalty adapts smoothly to varying curvature without introducing instability.
            \item For example:
            \[
            w_\text{curve} = \frac{\text{curvature}}{\text{curvature} + \gamma}.
            \]
            \item For dynamically changing curvature, an adaptive \(\gamma\) can be used:
            \[
            \gamma = \alpha \cdot \text{mean(curvature)},
            \]
            where:
            \begin{itemize}
                \item \(\alpha > 1\) is a scaling factor to adjust \(\gamma\) dynamically based on the curvature distribution.
                \item \(\text{mean(curvature)}\) represents the average curvature over a given segment of the path.
            \end{itemize}
            
            This approach ensures that \(\gamma\) adapts in real time, allowing the weighting function to remain sensitive to path complexity while maintaining numerical stability. By dynamically adjusting \(\gamma\), the steering penalty can balance responsiveness and smoothness across varying curvature scenarios.
        \end{itemize}
    \item Path-Aware Adjustments:
        \begin{itemize}
            \item Include path features, such as lane boundaries or obstacle positions, to refine the steering penalty in challenging scenarios.
    \end{itemize}
     \item Experimental Validation:
        \begin{itemize}
            \item Test the effect of curvature-weighted penalties on complex trajectories and evaluate the trade-off between trajectory smoothness and successful navigation through sharp turns.
    \end{itemize}
\end{enumerate}

Low-Speed Scenarios. At low speeds, steering changes have a reduced impact on stability and vehicle dynamics. Penalizing these changes equally across all speeds can lead to overly cautious behavior, especially in scenarios where quick adjustments are required to correct minor deviations.

To adapt the penalty, a speed-dependent scaling factor can be introduced:

\[
R_\text{steer} = -k \cdot |\Delta \theta_\text{steer}| \cdot (1 - w_\text{curve}) \cdot v_\text{scale},
\]

because for the low speed environment the speed already set to 1 m/s so \(v_\text{scale}\) set as 1, then the low speed scenarios formula is the same as sharp turns.

Comparison Of Steering Penalty Rewards:
\begin{figure}[h]
    \centering
    \includegraphics[width=1\linewidth]{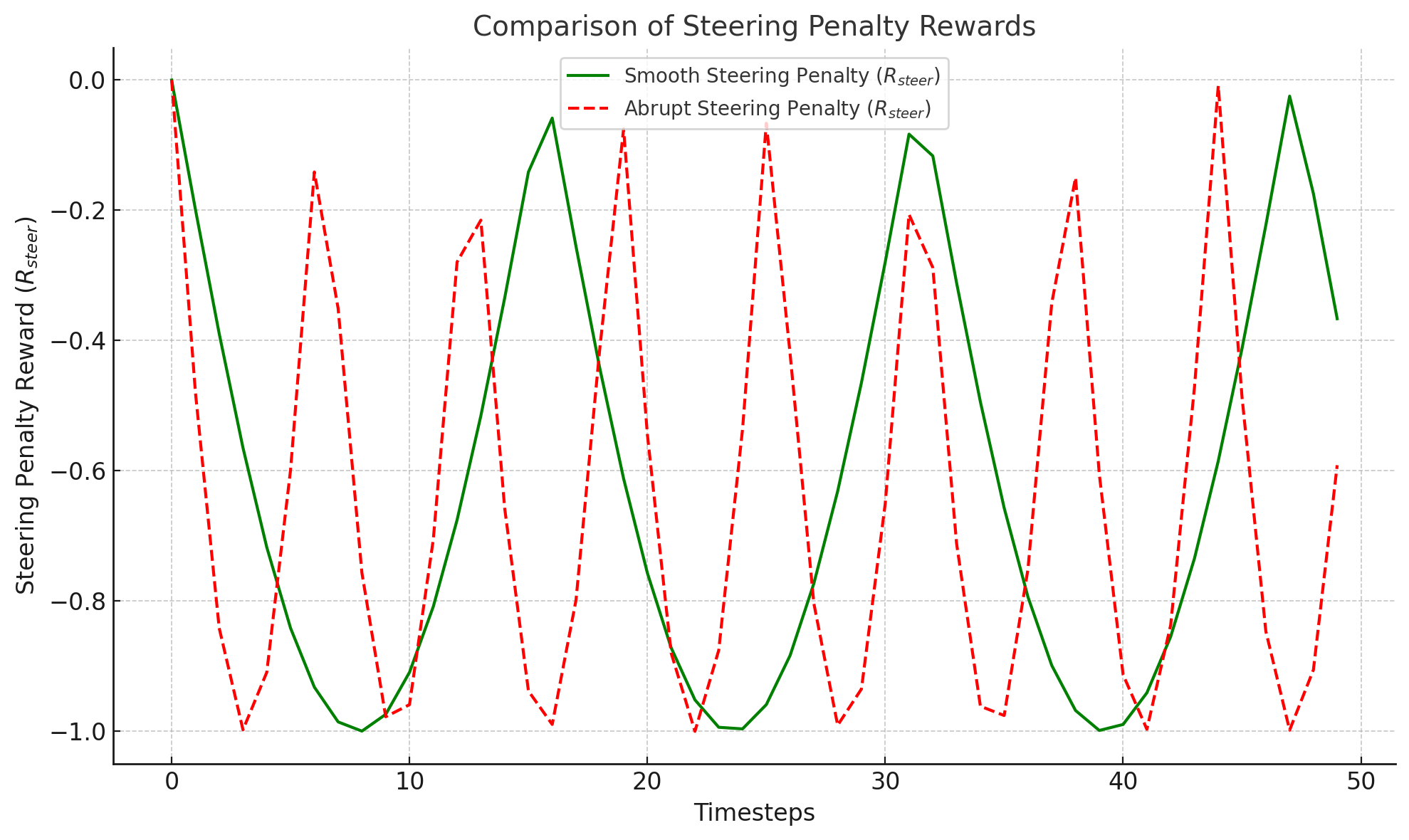}
    \caption{Steering Penalty Rewards}
    \label{fig:enter-label}
\end{figure}

\paragraph{Summary of Steering Penalty Comparison}
The generated plot compares the steering penalty rewards under two different scenarios: smooth steering and abrupt steering. The penalty rewards were calculated using the formula \(R_{\text{steer}} = -k \cdot \left| \Delta \theta_{\text{steer}} \right|\).

Key Observations:

\begin{enumerate}
     \item Smooth Steering:
        \begin{itemize}
            \item The penalty rewards remain relatively low and consistent across timesteps.
            \item This reflects a stable and controlled trajectory, where changes in the steering angle are minimal.
        \end{itemize}
    \item Abrupt Steering:
        \begin{itemize}
            \item The penalty rewards exhibit significant fluctuations, with higher penalties corresponding to larger steering angle changes.
            \item This indicates an unstable trajectory with frequent and substantial steering adjustments, which could compromise vehicle stability and passenger comfort.
        \end{itemize}
    \item Impact of \(k\):
        \begin{itemize}
            \item The proportionality constant \(k\) scales the penalty rewards uniformly across both scenarios, highlighting its role in controlling the overall penalty magnitude.
        \end{itemize}
\end{enumerate}

This Steering Delta Penalty can be integrated into the overall reward function for an autonomous driving agent, encouraging the vehicle to maintain smooth and controlled steering behavior while optimizing other factors such as speed and progress along the route.

\paragraph{}
Next, compare the formula \(R_{\text{steer}} = -k \cdot \left| \Delta \theta_{\text{steer}} \right|\) with \(R_\text{steer} = -k \cdot |\Delta \theta_\text{steer}| \cdot (1 - w_\text{curve})\), the result as below:

\begin{figure}[h]
    \centering
    \includegraphics[width=1.0\linewidth]{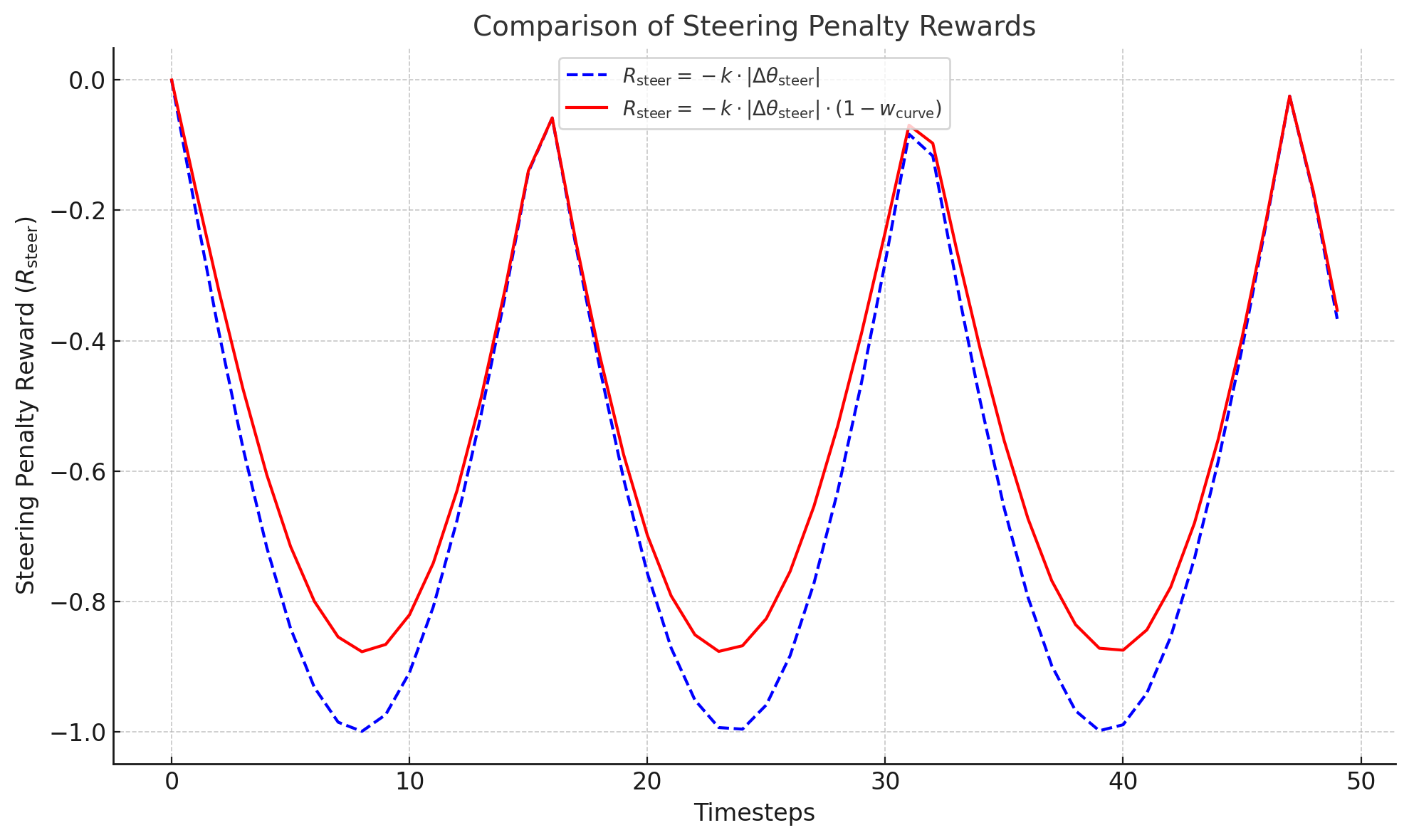}
    \caption{Comparison Of Steering Penalty Rewards}
    \label{fig:steer2}
\end{figure}

\newpage

Figure~\ref{fig:steer2} illustrates the comparison between two steering penalty formulations:
\begin{itemize}
    \item \textbf{Unweighted Penalty:} \( R_\text{steer} = -k \cdot |\Delta \theta_\text{steer}| \), where the penalty is directly proportional to the absolute change in the steering angle.
    \item \textbf{Weighted Penalty:} \( R_\text{steer} = -k \cdot |\Delta \theta_\text{steer}| \cdot (1 - w_\text{curve}) \), where the curvature-based weighting factor \( w_\text{curve} \) reduces the penalty in high-curvature scenarios.
\end{itemize}

\paragraph{Key Observations:}
\begin{enumerate}
    \item \textbf{Unweighted Penalty:} 
    The penalty remains constant for a given steering change, regardless of the path curvature. This results in higher penalties for necessary steering adjustments in sharp turns, potentially discouraging appropriate maneuvers.
    \item \textbf{Weighted Penalty:}
    The curvature-based weighting reduces penalties for large steering adjustments in high-curvature regions. This encourages more aggressive but necessary steering actions, improving trajectory adaptability in complex paths.
    \item \textbf{Smoothness:}
    The weighted penalty shows smoother transitions in penalties, reflecting its adaptability to varying curvature.
\end{enumerate}

\paragraph{Implications:}
The comparison highlights the importance of incorporating curvature-based weighting (\(w_\text{curve}\)) in steering penalties. While the unweighted penalty ensures stability on straight paths, the weighted penalty offers better flexibility in scenarios requiring sharp turns or dynamic maneuvers. The balance between these two approaches can be adjusted by tuning the curvature scaling factor (\(\gamma\)).

\subsection{Integration into Composite Reward Functions}

To optimize autonomous driving performance, steering penalties can be integrated with a velocity-based reward to form a composite reward function. This ensures that the agent is encouraged to maintain an optimal balance between speed and trajectory smoothness while adapting to path curvature.

\paragraph{Composite Reward Function},The composite reward function can be expressed as:
\[
R_\text{total} = w_\text{progress} \cdot R_\text{progress} + w_\text{steer} \cdot R_\text{steer} + w_\text{velocity} \cdot R_\text{velocity},
\]

\paragraph{where:}
\begin{itemize}
    \item \( R_\text{progress} \): Encourages progress along the track.
         \[
            R_\text{progress} = \frac{\Delta \text{Progress}}{\Delta L + \epsilon}.
        \]
    \item \( R_\text{steer} \): Discourages abrupt steering changes:
        \[
            R_\text{steer} = -k \cdot |\Delta \theta_\text{steer}| \cdot (1 - w_\text{curve}).
        \]
    \item \( R_\text{velocity} \): Encourages the agent to maintain the target velocity:
    \[
    R_\text{velocity} = e^{-\alpha \cdot |v_\text{actual} - v_\text{target}|}.
    \]
    \item \( w_\text{progress}, w_\text{steer}, w_\text{velocity} \): Weighting factors that balance the contributions of each reward component.
\end{itemize}

\paragraph{Benefits:}
\begin{enumerate}
    \item \textbf{Trajectory Smoothness:} \( R_\text{steer} \) reduces oscillations and ensures stability, with the weighted formulation allowing flexibility in high-curvature regions.
    \item \textbf{Speed Optimization:} \( R_\text{velocity} \) incentivizes the agent to maintain an optimal speed while balancing safety.
    \item \textbf{Progress Maximization:} \( R_\text{progress} \) ensures that the agent prioritizes forward motion and avoids unnecessary detours.
\end{enumerate}

\paragraph{Implementation:}
The weights \( w_\text{progress}, w_\text{steer}, w_\text{velocity} \) can be dynamically adjusted based on the track segment:
\begin{itemize}
    \item For straight segments, increase \( w_\text{velocity} \) to prioritize speed.
    \item For curved segments, increase \( w_\text{steer} \) to emphasize trajectory smoothness.
\end{itemize}

This integration allows the agent to balance trajectory smoothness, speed, and progress efficiently across diverse driving scenarios.

\section{EXPERIMENT}

\subsection{Experimental Validation}

For experimental validation, we employed the AWS DeepRacer Student League platform, which provides a simulated environment for autonomous driving tasks. The training parameters are summarized in Table~\ref{tab:training_parameters}.

\begin{table}[h]
    \centering
    \caption{Training Parameters}
    \label{tab:training_parameters}
    \resizebox{1.2\linewidth}{!}{
    \begin{tabular}{|l|l|}
        \hline
        \textbf{Parameter}         & \textbf{Value} \\ \hline
        Speed Range                & 0.1 m/s to 1.0 m/s \\ \hline
        Steering Angle Range       & -30° to 30° \\ \hline
        Action Space               & Continuous \\ \hline
        Training Duration          & Up to 10 hours \\ \hline
        Reward Function            & Composite(progress, velocity, steering penalties) \\ \hline
        Simulation Platform        & AWS DeepRacer Student League \\ \hline
    \end{tabular}
    }
\end{table}

\paragraph{Experimental Setup:}
The vehicle operates within a continuous action space, where the speed is constrained between 0.5 m/s and 1 m/s, and steering angles range from -30° to 30°. The composite reward function described earlier was utilized to balance progress, velocity optimization, and trajectory smoothness.

\paragraph{Training Strategy:}
\begin{itemize}
    \item \textbf{Duration:} The model was trained for a maximum of 10 hours to ensure convergence.
    \item \textbf{Scenarios:} Training covered various track conditions, including straight paths, high-curvature segments, and mixed terrain.
    \item \textbf{Hyperparameter Tuning:} We tuned reward function weights (\(w_\text{progress}, w_\text{steer}, w_\text{velocity}\)) to optimize performance for different scenarios.
\end{itemize}

\paragraph{Results:}
The trained model exhibited robust performance across diverse driving scenarios. Key findings include:
\begin{enumerate}
    \item The weighted composite reward function significantly improved trajectory smoothness on high-curvature tracks compared to the unweighted version.
    \item Velocity optimization resulted in efficient completion times while maintaining stability.
    \item The integration of progress, velocity, and steering penalties enabled the vehicle to adapt effectively to both straight and curved paths.
\end{enumerate}

These results demonstrate the effectiveness of the composite reward function and the training methodology in achieving robust and efficient autonomous driving performance.

\paragraph{}
Participation in the AWS DeepRacer Student League took place from March to October, spanning several months of intense competition. Through consistent performance and strategic optimization, the team successfully advanced to the final round. In the highly competitive final, the model emerged victorious, securing first place.

\paragraph{}
As an integral member of the team, Elena Wu, currently pursuing her university studies in the United States, played a crucial role in the success. Her dedication, expertise, and strategic contributions were instrumental in navigating the challenges of the competition. This year, Elena achieved a remarkable milestone by winning the first-place title in the AWS DeepRacer Student League Grand Finale, showcasing her exceptional skills and commitment to excellence.

\begin{figure}[h]
    \centering
    \includegraphics[width=1\linewidth]{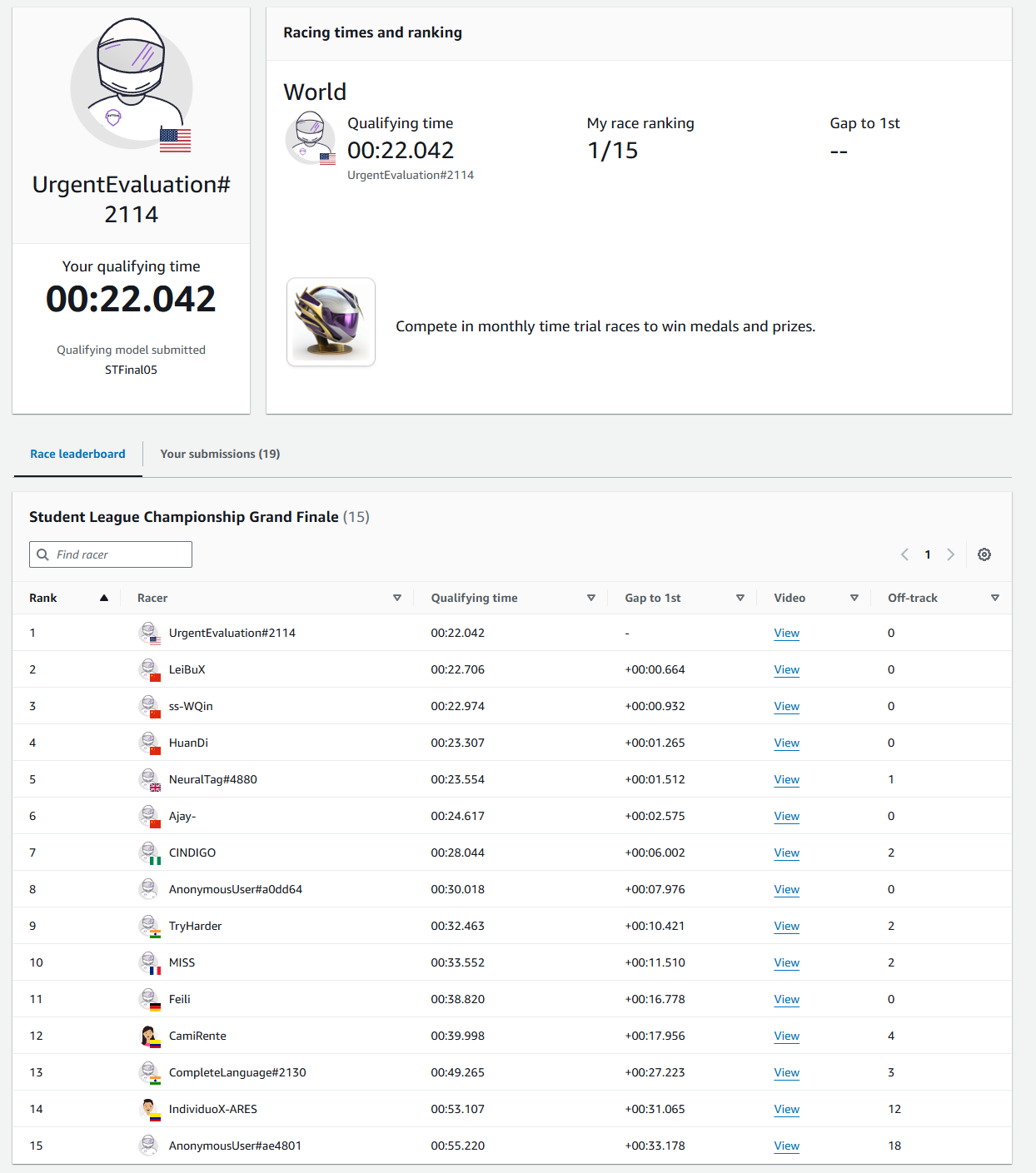}
    \caption{Deepracer student League Final}
    \label{fig:enter-label}
\end{figure}

\newpage
\bibliographystyle{plain} 
\bibliography{references} 

\end{document}